\documentclass{article}

\usepackage[utf8]{inputenc} 
\usepackage[T1]{fontenc}    
\usepackage{hyperref}       
\usepackage{url}            
\usepackage{booktabs}       
\usepackage{amsfonts}       
\usepackage{nicefrac}       
\usepackage{microtype}      
\usepackage{xcolor}         
\usepackage{graphicx}       
\usepackage{amsmath}
\usepackage{multirow}
\usepackage{float}
\usepackage{array}
\usepackage{pifont}
\usepackage{wrapfig}
\usepackage{adjustbox}
\usepackage{caption}
\usepackage{makecell}
\usepackage{authblk}

\title{\Large RS-GPT4V: A Unified Multimodal Instruction-Following \\ Dataset for Remote Sensing Image Understanding}

\author{Linrui Xu, Ling Zhao, Wang Guo, Qiujun Li, Kewang Long, Kaiqi Zou, Yuhan Wang, Haifeng Li\thanks{Corresponding author email: liaifeng@csu.edu.cn}}

\affil{Central South University, Changsha, China}

\begin{document}
    
\maketitle

\begin{abstract}
    The remote sensing image (RSI) intelligence understanding model is undergoing a new profound paradigm shift which has been promoted by multi-modal large language model (MLLM), i.e. from the paradigm learning a domain model (LaDM) shifts to paradigm learning a pre-trained general foundation model followed by an adaptive domain model (LaGD). Under the new LaGD paradigm, the old datasets, such as RSI-CD, DOTA, which have led to advances in RSI intelligence understanding in the last decade, are no longer suitable for fire-new tasks. We argued that a new dataset must be carefully designed to lighten tasks in the new paradigm with the following features: (1) Generalization: training model to learn shared knowledge among tasks and to adapt to different tasks; (2) Understanding complex scenes: training model to understand the fine-grained attribute of the objects of interest, and to be able to describe the scene with natural language with detailed; (3) Reasoning: training model to be able to realize high-level visual reasoning. In this paper, we designed a high-quality, diversified, and unified multimodal instruction-following dataset for RSI understanding produced by GPT-4V and existing datasets, which we called RS-GPT4V. To achieve generalization, RS-GPT4V used a (Question, Answer) which was deduced from GPT-4V via instruction-following to unify the tasks such as captioning, localization; To achieve an understanding of a complex scene, RS-GPT4V proposed a hierarchical instruction description with local strategy in which the fine-grained attributes of the objects and their spatial relationships are described and global strategy in which all the local information are integrated to yield detailed instruction descript; To achieve reasoning, RS-GPT4V designed multiple-turn (Question, Answer) pair to provide the reasoning ability for a model. The empirical results show that the fine-tuned MLLMs by RS-GPT4V can describe fine-grained information, and implicit knowledge in multiple complex remote sensing scenarios, and reason better than existing datasets. The source code and dataset can be visited at: https://github.com/GeoX-Lab/RS-GPT4V.
\end{abstract}

\section{Introduction}
    In the LaDM paradigm, the datasets were designed for research tasks such as scene classification\cite{xia2017aid}, object detection\cite{ren2015faster}, image captioning\cite{karpathy2015deep}, and visual question answering\cite{li2019visualbert}. Research typically relies on designing and training models separately for each task, neglecting the potential commonalities and knowledge sharing between different tasks and datasets. Remote sensing interpretation datasets primarily consist of images and annotations, where images are consistent entity samples, while annotations are diverse, including labels, bounding boxes, and text. Additionally, existing datasets usually preset a limited number of scenes and target categories, mainly examining the model's ability to recognize these specific categories, overlooking whether the model can deeply understand and reason about the complex relationships between various scenes and targets.

    In the LaGD paradigm, significant achievements in language understanding and reasoning capabilities have been made in recent years, driven by the rapid development of large language models (LLMs) such as GPT-4\cite{achiam2023gpt} and LLaMA2\cite{touvron2023llama}. To further harness the potential of LLMs, researchers have expanded their capabilities by integrating visual perception modules, thus forming MLLMs capable of understanding complex image scenes and performing visual reasoning. The training paradigm for MLLMs typically follows a pre-training and fine-tuning model, where visual language instruction fine-tuning plays a crucial role throughout the training process. When MLLMs undergo Supervised Fine-Tuning (SFT) with appropriate visual language instruction-following datasets, they demonstrate powerful functionalities. They can not only perceive, understand, and process visual information but also enhance their performance in executing complex task instructions. Visual language instruction fine-tuning provides the model with the ability to comprehend and follow visual content-related instructions, enabling it to generate responses that meet user interaction needs.
    
    However, the existing remote sensing datasets and benchmarks present three challenges: (1) The diversity of annotations in remote sensing datasets limits the generalization capability of models. The inconsistency in annotation modalities makes it difficult for models to adapt to different tasks. (2) Remote sensing annotation data are insufficient to accurately describe the fine-grained attributes of objects in the area of interest and the structural information between these fine-grained attributes. (3) Existing methods can only perform one-loop recognition (OLR) and are unable to achieve multi-turn reasoning (MTR) to discover implicit knowledge.
    
    Despite MLLMs demonstrating excellent performance in various fields, the diversity of annotations in existing remote sensing datasets limits their application. Moreover, current annotation data have limitations in providing detailed descriptions of RSI targets, particularly regarding fine-grained attributes, spatial relationships, and background knowledge, as shown in Figure \ref{fig:1}. Therefore, this paper proposes a Unified RS Multimodal Instruction-Following Dataset (RS-GPT4V). This dataset uses the text modality to express different annotations as a unified foundation. By converting multiple tasks into language-understanding tasks, data unification can be achieved, further realizing task unification. RS-GPT4V aims to cover a wide range of scenarios and target categories and integrate various visual language tasks. The dataset adopts a unified (Question, Answer) format, supporting tasks such as image description, visual question answering, complex scene understanding, visual reasoning, and task planning.

    RS-GPT4V is constructed through two key methods: Instruction-Annotation Adaption and Instruction-Response Generation. Instruction-Annotation Adaption converts existing visual language tasks into (Question, Answer) pairs using instruction templates. Instruction-Response Generation utilizes system prompts and advanced GPT-4V models to generate (Question, Answer) pairs based on existing annotation data. By using RS-GPT4V for supervised fine-tuning of MLLMs, the model can understand the relationships between objects in complex scenes and uncover implicit knowledge. For example, using contextual visual information of a ship's wake to infer whether the target is stationary or moving.

    Our contributions are listed as follows:


\begin{itemize}

    \item 
    We achieve a new high-quality, diversified, and unified multimodal instruction-following dataset by (Question, Answer) pair as the uniform form for RSI understanding produced by GPT4V and existing datasets, called RS-GPT4V for the new paradigm. The RS-GPT4V could be used to train and test models' capabilities in generalization, understanding complex scenes, and reasoning.
    
    \item 
    The designed some principles, such as such as uniformity, diversity, accuracy, and richness, to design a dataset. These principles are general and could serve as guidance for the LaGD paradigm in the future. 

    \item 
    The empirical results show that the fine-tuned MLLMs by RS-GPT4V can describe fine-grained information, implicit knowledge in multiple complex remote sensing scenarios, and reason better than existing datasets. We will release the dataset and code.
    
\end{itemize}


\section{Related work}
    \label{gen_inst}


\textbf{Remote Sensing Visual Language Datasets.} 
    RSI datasets are typically categorized by task types, such as classification\cite{xia2017aid}, segmentation\cite{wang2023samrs}, detection\cite{sun2021fair1m}, and change detection\cite{ji2024changenet}. The primary distinction among these datasets lies in their annotation methods, with classification tasks using class indices and detection tasks using bounding boxes. With the advancement of vision-language models\cite{radford2021learning}, researchers have developed remote sensing vision-language datasets, such as RS5M\cite{Long2021DiRS}, SkyScrip\cite{wang2023skyscript}, and LHRS-Align\cite{muhtar2024lhrsbot}, by integrating RSI with language descriptions. These datasets aim to achieve a unified annotation approach using language modalities, thereby enabling the development of vision-language models capable of handling various downstream tasks. While these datasets have demonstrated significant advantages in tasks like image captioning and scene classification, they have yet to achieve complete task unification, as models trained on these datasets are typically limited to specific tasks within the dataset. This work posits that the annotations for different tasks are essentially information carriers that can be expressed through language modalities. By learning the mapping relationship between images and language annotations, vision-language models have the potential to handle diverse tasks in a unified manner.

\textbf{Remote Sensing Multimodal Instruction-Following Datasets.}
     Although using language modalities to unify annotations across different tasks can facilitate task unification within datasets, enabling vision-language models to select downstream tasks autonomously remains a challenge. To address this issue, researchers have started incorporating task instructions into model training. Specifically, instruction-guided multimodal datasets introduce task-specific texts such as "describe the image," "locate objects in the image," or "answer the question briefly" alongside visual and language inputs. These texts guide the model in understanding the task type and its corresponding requirements. Through this common interface, the same model can be fine-tuned for various visual tasks, resulting in a general-purpose multimodal model capable of accepting any language and visual inputs and solving diverse visual tasks.\cite{huang2023visual}

\begin{figure}[ht]
  \centering
  \includegraphics[width=10cm]{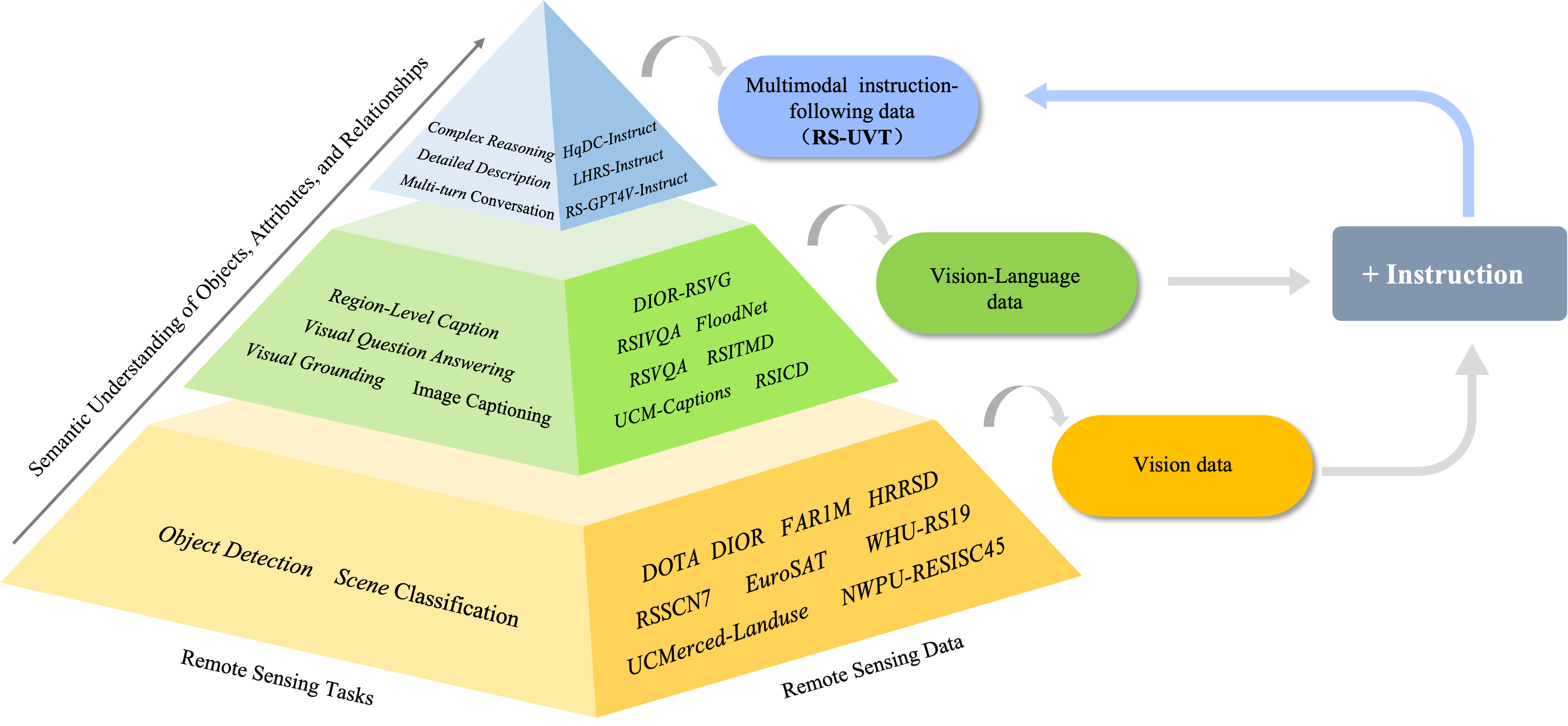}
  \caption{\textbf{Evolution of Remote Sensing Tasks and Data.} The figure illustrates the progression from early remote sensing tasks using purely visual data to advanced tasks using visual language data. By adding instructions, these data can be transformed into multimodal instruction-following datasets.}
  \label{fig:1}
\end{figure}

    Currently, notable multimodal instruction-following datasets in remote sensing include MMRS-1M\cite{zhang2024earthgpt}, RS-instructions\cite{rs16091477}, and RS-Specialized-Instruct\cite{pang2024h2rsvlm}. As shown in Figure \ref{fig:1}, These datasets enhance existing vision-language remote sensing datasets like UCM-Captions\cite{qu2016deep}, RSVQA\cite{lobry2020rsvqa}, and DIOR\cite{Li_2020} by adding instructional text, merging them into new multimodal Instruction-Following datasets. Additionally, datasets like HqDC-Instruct\cite{pang2024h2rsvlm} and LHRS-Instruct\cite{muhtar2024lhrsbot} leverage large language models to construct new task instructions for multi-turn Conversations, detailed descriptions, and complex reasoning. However, these datasets often lack systematic guiding principles. For instance, MMRS-1M\cite{zhang2024earthgpt} lacks complexity, making it difficult for models to handle complex semantic reasoning tasks. GeoChat's\cite{kuckreja2023geochat} data generation process lacks supervision, leading to insufficient correctness and a higher likelihood of contextually inconsistent predictions. Therefore, this work aims to redesign the principles for constructing remote sensing instruction fine-tuning datasets and, under these guidelines, create more unified, rich, and correct remote sensing multimodal instruction-following datasets.

\section{RS-GPT4V Dataset}
    \label{headings}    

\subsection{Design Principles and Characteristics}
    \label{3.1}

\begin{figure}[ht]
  \centering
  \includegraphics[width=13cm]{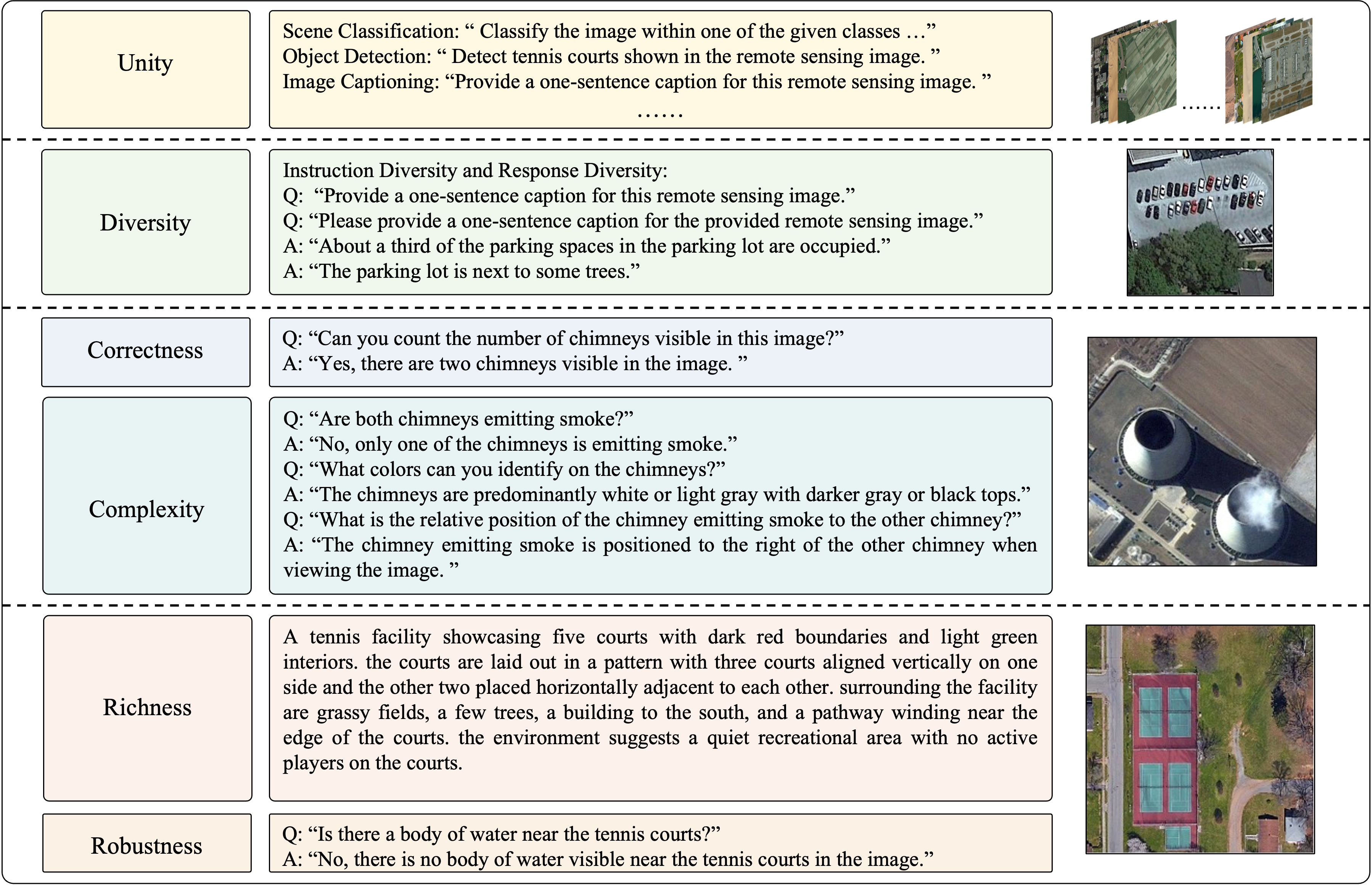}
  \caption{Design Principles and Characteristics of the RS-GPT4V Dataset.}
  \label{fig:2}
\end{figure}

    In constructing the RS-GPT4V Dataset, we adhered to several key design principles, as illustrated in Figure \ref{fig:2}, to ensure that the remote sensing instruction-following dataset is not only of high quality but also broadly applicable to various visual and visual-language tasks. The design principles are summarized as follows:

    \textbf{Unity:} Our dataset integrates multiple visual and visual-language tasks, such as image description, visual question answering, and visual dialogue. By adopting a unified data annotation, we ensure consistency and standardization, allowing a single model to be trained and evaluated across multiple tasks.
    
    \textbf{Diversity: }To accommodate different remote sensing scenarios and research needs, the instruction-following dataset includes images from various scenes along with their corresponding instructions and questions. This diversity is reflected not only in the visual content but also in the types of tasks and model responses.\cite{zhang2023instruction}
    
    \textbf{Correctness: }The matching of visual information with textual content must be precise. All images and text descriptions in the RS-GPT4V Dataset have been manually corrected to ensure error-free image-text question answering.\cite{li2023vision}
    
    \textbf{Richness:} In addition to basic visual and textual data, we have included supplementary background knowledge, such as the scientific background of related objects. This rich information helps the model understand and interpret the complex elements and relationships within the visual content better.

    \textbf{Complexity:} Our dataset design includes multi-layered tasks and challenges, particularly in reasoning about object attributes and relationships. By designing tasks that require complex logic and reasoning capabilities, we enhance the model's ability to handle intricate scenarios.\cite{li2023vision,zhang2023instruction}
    
    \textbf{Robustness:} The instruction-following dataset includes additional spatial negative samples to enhance the model's robustness, ensuring stable performance across various remote sensing scenarios and reducing hallucinations.\cite{you2023ferret}

\subsection{Principles-Driven Pipeline for RS-GPT4V Dataset Construction}

\begin{figure}[ht]
  \centering
  \includegraphics[width=14cm]{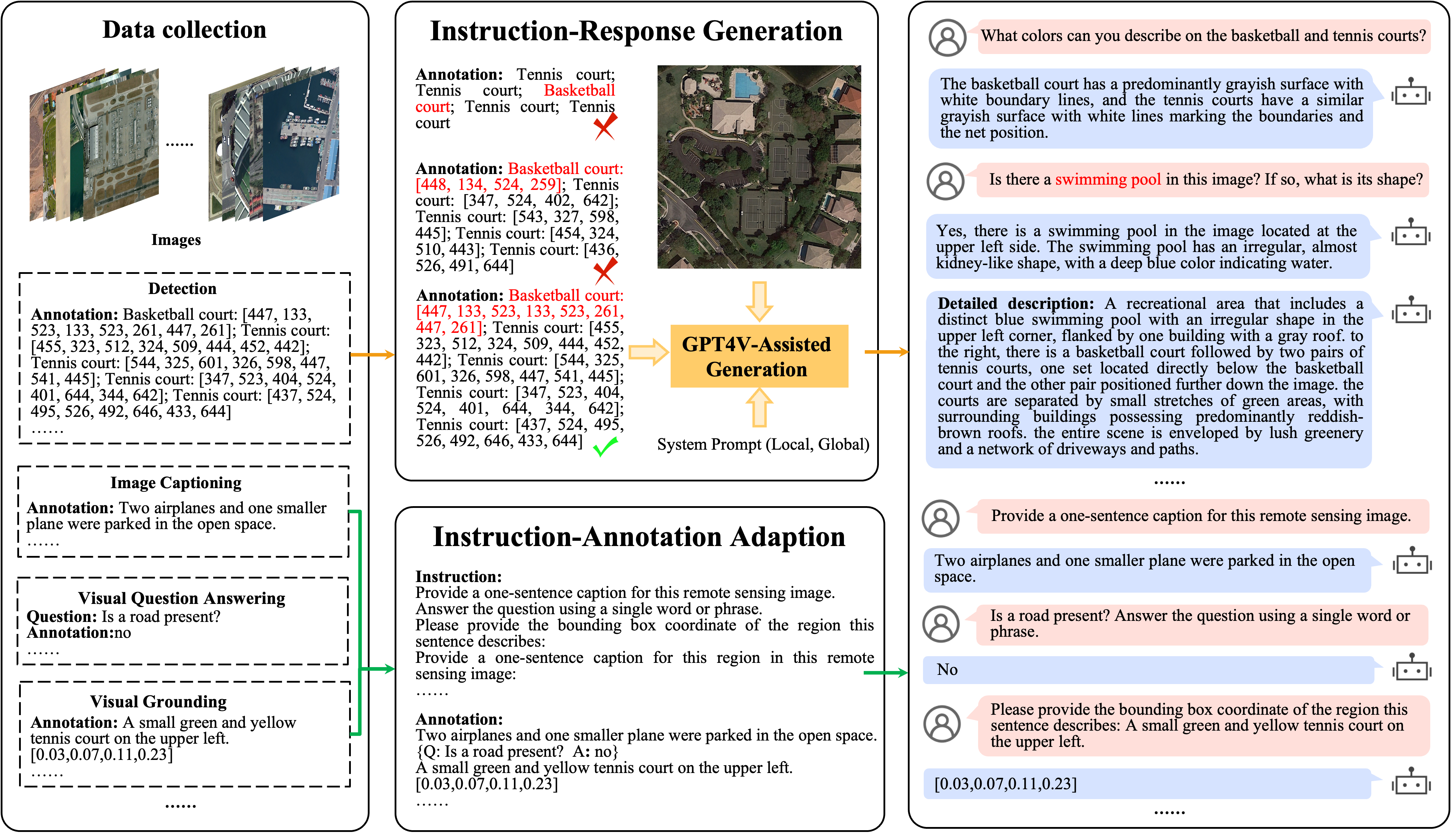}
  \caption{\textbf{Principles-Driven Pipeline for RS-GPT4V Dataset Construction.} The construction process of the RS-GPT4V dataset follows design principles for unified remote sensing tasks. The pipeline includes data collection from various remote sensing datasets, instruction-response generation with detailed descriptions, complex reasoning, and multi-turn conversation, and instruction-annotation adaption to create a comprehensive and accurate dataset.}
  \label{fig:3}
\end{figure}

    During the construction of the RS-GPT4V Dataset, we adhered to the design principles\cite{bsharat2023principled,li2023vision} outlined in Section \ref{3.1} to ensure the dataset's quality, accuracy, and diversity, as shown in Figure \ref{fig:3}. The following describes the detailed pipeline for dataset construction and how these principles were applied within the pipeline.

    \textbf{Data Collection:} We extracted data from multiple well-recognized remote sensing visual language datasets. These datasets cover a wide range of scenes, from natural landscapes to urban environments, providing a rich source of RSI and related annotations. Although these datasets serve their respective purposes, they often lack the depth and complexity required to address more complex challenges in multi-turn interactions. By collecting data from these diverse sources, each dataset was carefully selected to cover different geographical regions and environmental conditions, ensuring data diversity. For example, some datasets contain high-resolution images of urban areas, while others focus on natural landscapes or agricultural regions. To construct a dataset capable of complex reasoning, we selected the DIOR\cite{Li_2020} dataset with coordinate annotations, which have been manually labeled to provide accurate coordinate hints. In contrast, classification datasets containing only a single category label are not suitable for GPT-4V due to the difficulty and high error rate in image understanding.
    
    \textbf{Instruction-Response Generation:} As shown in the diagram, we adopted a hierarchical prompting instruction description generation method. First, we obtained fine-grained information about objects at a local level and combined this information with RSIs to systematically generate detailed instruction descriptions. To ensure that the generated instructions are detailed and accurate, each instruction and its corresponding response must accurately reflect the image content and possess sufficient complexity and diversity. Specifically, the coordinate information from the DIOR dataset was used to identify different types of objects in the image, such as tennis courts and basketball courts. To avoid the influence of non-continuous objects on GPT-4V's\cite{achiam2023gpt} counting ability, we ensured that the sequence of labeled objects was continuous. For example, the sequence would be "tennis court, tennis court, basketball court" rather than "tennis court, basketball court, tennis court". To enhance GPT-4V's spatial understanding ability, we used rotated bounding boxes to represent the positions of objects. Using rotated bounding boxes instead of standard bounding boxes helps avoid coordinate overlap, enabling the model to better understand the fine-grained attributes and spatial relationships of objects. Meanwhile, the questions generated by GPT-4V should be sufficiently challenging, covering complex reasoning, world knowledge, explanatory answers, and multi-turn dialogues. For example, "Please describe the basketball court in the image and its color," and "How many tennis courts are there in the image?". Combining the complex and diverse instruction information obtained at the local level with the RSIs, GPT-4V can generate more accurate and detailed responses. For example, identifying a basketball court in the image and generating a detailed description of its location, color, and surrounding environment.

    \textbf{Instruction-Annotation Adaption:} Adapting annotations from visual language datasets to instruction datasets is a crucial part of constructing the RS-GPT4V Dataset. By adding specific instructions, we transformed existing visual language datasets into new instruction datasets. For example, in image description tasks, the original dataset might only contain simple descriptions, and by adding instructions such as "Provide a one-sentence caption for this RSI," the dataset becomes more suitable for complex instruction-response tasks. Through this method, the RS-GPT4V Dataset not only ensures task diversity and accuracy at the data collection stage but also enhances dataset unity through meticulous instruction generation and annotation transformation steps.

\subsection{Overview of the RS-GPT4V Dataset}

     As shown in Table \ref{tab:1}. The RS-GPT4V dataset comprises multiple tasks with specific RSIs and instruction annotations for tasks such as image captioning, visual question answering, visual grounding, and region-level descriptions. Key datasets include NWPU-Captions\cite{cheng2022nwpu}, RSICD\cite{lu2017exploring}, RSITMD\cite{RSITMD}, Sydney-Captions\cite{qu2016deep}, UCM-Captions\cite{qu2016deep}, RSVQA-LR\cite{lobry2020rsvqa}, RSVQA-HR\cite{lobry2020rsvqa}, FloodNet\cite{rahnemoonfar2021floodnet}, RSIVQA\cite{RSIVQA}, and DIOR-RSVG\cite{zhan2023rsvg}, et al. Overall, the dataset provides 91,937 training images with 991,206 question-answer pairs and 15,999 test images with 258,419 question-answer pairs, supporting detailed descriptions and complex reasoning capabilities across various visual and visual-language tasks.

    \begin{table}[ht]
    \vspace{-1mm}
    \caption{Details of the RS-GPT4V Dataset: Images and QA Pairs for Training and Testing}   
    \scriptsize 
    \label{tab:1}
    \setlength{\tabcolsep}{6pt} 
        \begin{tabular}{cccccc} 
            \toprule
            Task                     & Data Source         & Train images & Train QA Pairs & Test images & Test QA Pairs \\
            \midrule
            \multirow{5}{*}{Image Captioning} 
                                     & NWPU-Captions       & 25200        & 125894            & 3150        & 1093             \\
                                     & RSICD               & 8734         & 17813             & 1093        & 1093             \\
                                     & RSITMD              & 4291         & 20096             & -           & -                \\
                                     & Sydney-Captions     & 497          & 2294              & 58          & 58               \\
                                     & UCM-Captions        & 1680         & 7999              & 210         & 210              \\
            \multirow{4}{*}{Visual Question Answering}
                                     & RSVQA-LR           & 572          & 57223             & 100         & 10004            \\
                                     & RSVQA-HR           & 6251         & 625340            & 2226        & 222684           \\
                                     & FloodNet            & 1448         & 4511              & -           & -                \\
                                     & RSIVQA              & 5401         & 19218             & -           & -                \\
            Visual Grounding         & DIOR-RSVG           & 9466         & 19643             & 7936        & 18677            \\
            Region-level Captioning  & DIOR-RSVG           & 9466         & 19643             & -           & -                \\
            Multi-turn Conversation  & \textbf{RS-GPT4V-Instruct}   & 9466         & 62067             & 613         & 3987             \\
            Detailed Description     & \textbf{RS-GPT4V-Instruct}   & 9465         & 9465              & 613         & 613              \\
            \bottomrule
            Total                    & -                   & 91937        & \textbf{991206}            & 15999       & \textbf{258419}           \\
        \end{tabular}
        \vspace{-1mm}
\end{table}
    
    The RS-GPT4V dataset comprises a total of 91,937 training images and 991,206 training instruction-answer pairs, with 15,999 images and 258,419 instruction-answer pairs in the test set. By integrating these datasets, the RS-GPT4V dataset features detailed annotations and complex reasoning capabilities. The structure of this dataset supports a variety of tasks, ranging from image description to multi-turn dialogues and detailed descriptions.

\section{Experiments}
    \label{others}

\subsection{Experimental Setup}

    \label{Experimental}
    To evaluate the impact of different fine-tuning strategies on the RS-GPT4V dataset, we employed three fine-tuning methods to perform supervised fine-tuning based on the LLaVA-1.5-7B \cite{liu2024visual} model: Full-Parameter Fine-Tuning (Full-FT), LoRA Fine-Tuning, and MoE-LoRA Fine-Tuning. The objective of these strategies is to compare the efficacy and performance of various fine-tuning methods in handling complex remote sensing tasks. All experiments were conducted on four NVIDIA A800-80G GPUs, with a configuration that includes a global batch size of 64, an initial learning rate of 2e-4, and a total of 14,666 training steps, corresponding to 1 epoch. Additionally, the rank is set to 128, and the number of experts is 4.

\subsection{Benchmarks for RS-GPT4V}

    \textbf{Full-FT:} In the full-parameter fine-tuning strategy, we adjust all parameters of the LLaVA-1.5 model to fully adapt to the specific characteristics of RSIs. This comprehensive adjustment strategy optimizes all trainable parameters, enabling the model to handle complex remote sensing tasks more effectively.

    \textbf{LoRA:} The LoRA\cite{hu2021lora} fine-tuning strategy aims to improve the model's learning efficiency and reduce computational resource requirements by applying low-rank approximation optimization to specific parameters within the model. This strategy optimizes the structure of certain parameters, significantly reducing resource consumption and enabling training to be completed in a shorter time frame.

    \textbf{MoE-LoRA:} We employed the MoE-LoRA\cite{liu2023moelora} method, which combines the Mixture of Experts (MoE) and Low-Rank Adaptation (LoRA) approaches. This method leverages the advantages of MoE in multi-task learning and the efficiency of LoRA in parameter fine-tuning. By designing multiple experts, each consisting of a pair of low-rank matrices, this approach maintains computational efficiency while significantly enhancing the model's capability to handle multiple tasks.

\begin{figure}[ht]
  \centering
  \includegraphics[width=1\textwidth]{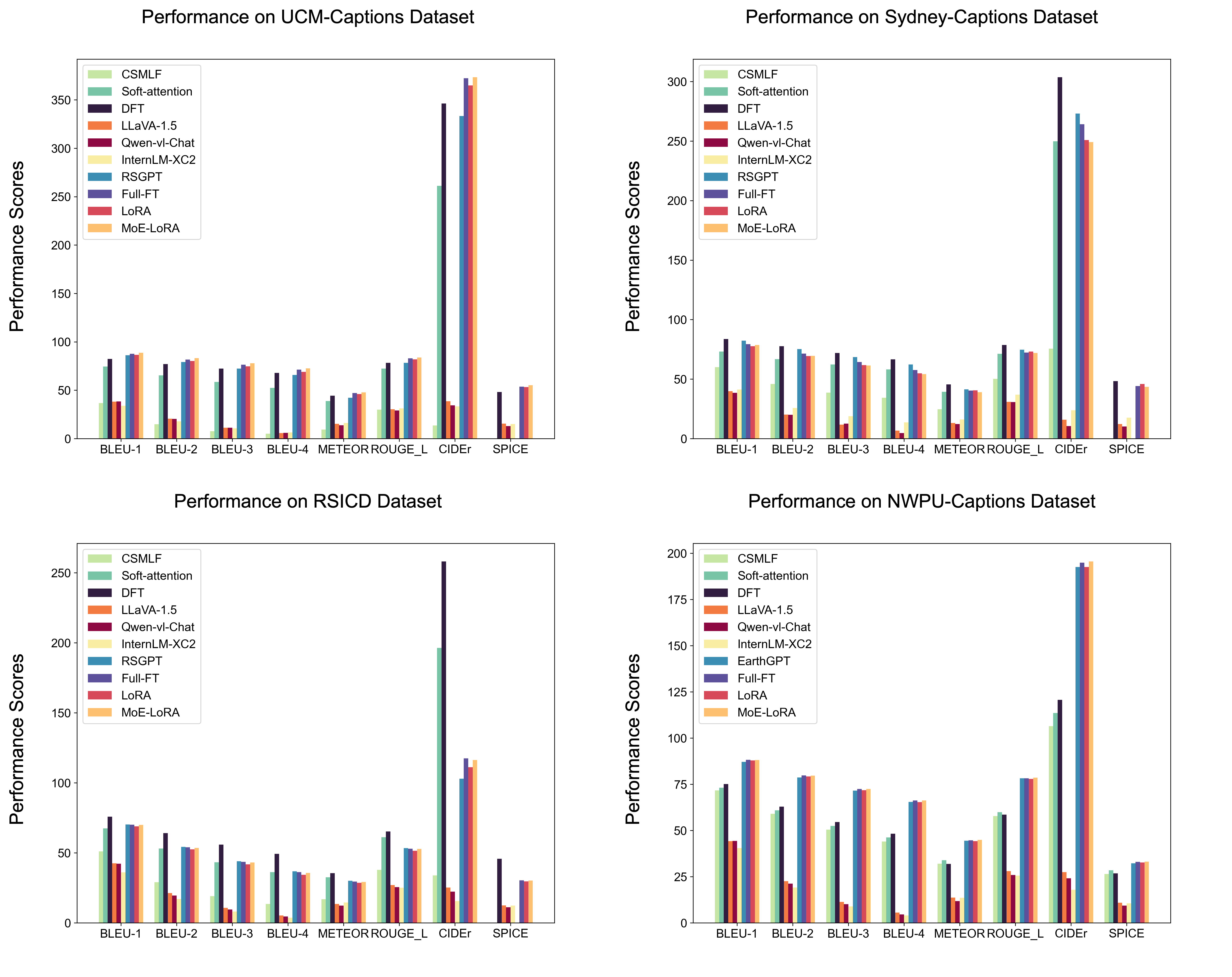}
  \caption{\textbf{Performance on Image Captioning Tasks Across Four Remote Sensing Datasets.} Comparison of our benchmark with current state-of-the-art methods on UCM-Captions, RSICD, Sydney-Captions, and NWPU-Captions datasets. Metrics include BLEU-1\cite{papineni2002bleu}, BLEU-2, BLEU-3, BLEU-4 (measure of $n$-gram precision), METEOR\cite{banerjee-lavie-2005-meteor} (measure of precision, recall, and $F_1$ score), ROUGE\_L\cite{ROUGE_L} (measure of longest common subsequence), CIDEr\cite{cider} (measure of consensus-based evaluation), and SPICE\cite{anderson2016spice} (measure of semantic propositional content).}
  \label{figure_5}
\end{figure}

\subsection{Experiments and Results}
\subsubsection{Image Captioning Task}

    In our experimental analysis, we evaluated the performance of three benchmark models (Full-FT, LoRA, MoE-LoRA) fine-tuned using RS-GPT4V on four different remote sensing datasets (UCM-Captions\cite{qu2016deep}, RSICD\cite{lu2017exploring}, Sydney-Captions\cite{qu2016deep}, NWPU-Captions\cite{cheng2022nwpu}). As shown in \ref{figure_5}, the fine-tuned benchmark models generally demonstrated significant performance improvements, providing accurate and detailed descriptions of RSIs. However, the performance on the Sydney-Captions dataset was not as strong as on the other datasets. This discrepancy is primarily due to the relatively small scale of the Sydney-Captions dataset, which resulted in the models being unable to fully capture the specific characteristics of this dataset during training. Nonetheless, the performance of the fine-tuned models on the other datasets still indicates that the RS-GPT4V fine-tuning strategy has significant potential and utility in enhancing the automatic parsing and description of RSI.


\subsubsection{Visual Question Answering Task}

\begin{figure}[ht]
  \centering
  \includegraphics[width=1\textwidth]{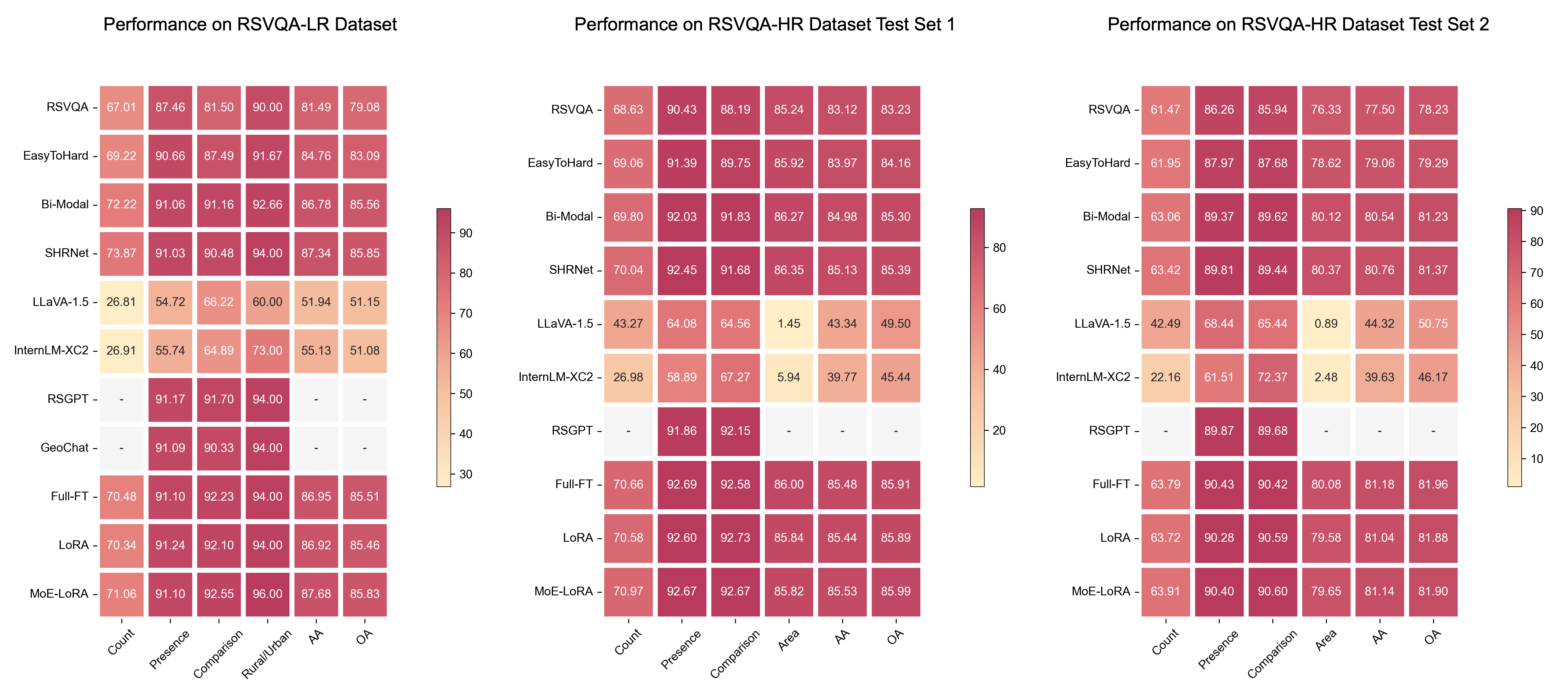}
  \caption{\textbf{Accuracy of Different Models on RSVQA-LR and RSVQA-HR Datasets.} Metrics include Count, Presence, Comparison, Rural/Urban, the average accuracy (AA), and the overall accuracy (OA).}
  \label{figure_6}
\end{figure}

    In the performance analysis of the RSVQA-LR and RSVQA-HR datasets (including Test set 1 and Test set 2), as shown in Figure \ref{figure_6}, the three benchmark models fine-tuned with RS-GPT4V demonstrated significant advantages in multiple key metrics. Although the Count metric in RSVQA-LR was slightly inferior to Bi-Modal\cite{bazi2022bi} and SHRNet\cite{zhang2023spatial} models, the MoE-LoRA benchmark performed exceptionally well across most evaluation metrics, particularly excelling in Presence, Comparison, and Rural/Urban. In both test sets of RSVQA-HR, all three models generally outperformed other methods, with MoE-LoRA's strong performance in high-resolution image analysis being particularly noteworthy. However, compared to Test set 1, the accuracy of all models decreased in Test set 2, potentially reflecting increased test set difficulty or changes in data distribution. Overall, despite the subpar performance in the Count metric, the excellent performance in Area and other advanced visual understanding tasks validates the effectiveness of the RS-GPT4V fine-tuning strategy in enhancing RSI understanding capabilities.

\subsubsection{Visual Grounding Task}

    In the analysis of the visual grounding task results, as shown in Table\ref{table_3}, the performance of different models on the Pr@0.5 metric revealed significant differences. Overall, the three benchmark models fine-tuned with RS-GPT4V, namely Full-FT, LoRA, and MoE-LoRA, demonstrated superior performance in this task, significantly outperforming the Qwen-vl-Chat\cite{bai2023qwen} and LLaVA-1.5\cite{liu2023improved} models. General MLLMs showed poor performance in detection tasks, while the models fine-tuned with RS-GPT4V exhibited strong capabilities in visual grounding tasks. These results validate the efficacy and robustness of the RS-GPT4V fine-tuning strategy in enhancing RSI visual grounding tasks.

\begin{table}[ht]
\setlength{\tabcolsep}{3pt} 
\begin{minipage}{0.35\linewidth}
  \centering
  \caption{\textbf{Performance of Models on DIOR-RSVG.} Compares the accuracy (Pr@0.5) of various models on the DIOR-RSVG dataset, using the DIOR dataset split for training and testing.}
  \resizebox{\linewidth}{!}{
    \begin{tabular}{@{}lc@{}}
        \toprule
        Method       & Accuracy@0.5 \\
        \midrule
        Qwen-vl-Chat & 25.05        \\
        LLaVA-1.5    & 9.52         \\
        Full-FT      & 36.31        \\
        LoRA         & 33.15        \\
        MoE-LoRA     & \textbf{37.86}        \\
        \bottomrule
    \end{tabular}
  }
  \label{table_3}
\end{minipage}
\hspace{6mm}
\begin{minipage}{0.6\linewidth}
  \centering
  \caption{\textbf{GPT-4V Performance Evaluation.} GPT-4V scores the performance of various models on the RS-GPT4V-Instruct dataset, focusing on complex conversation and detailed description. Scores are restricted between 1 and 10, with higher scores awarded for answers containing more accurate information.}
  \resizebox{\linewidth}{!}{
     \begin{tabular}{@{}lccc@{}}
        \toprule
        Method        & \makecell[c]{complex reasoning \\ \& conversation} & detailed description & All Scores \\
        \midrule
        LLaVA-1.5     & 5.21                 & 5.088                & 5.194      \\
        Qwen-vl-Chat  & 2.648                & 2.282                & 2.599      \\
        InternLM-XC2  & 5.312                & 4.392                & 5.189      \\
        Full-FT       & \textbf{6.27}        & \textbf{6.53 }                & \textbf{6.304  }    \\
        LoRA          & 6.061                & 6.374                & 6.103      \\
        MoE-LoRA      & 6.108                & 6.468                & 6.156      \\
        \bottomrule
    \end{tabular}
  }
  \label{table_4}
\end{minipage}
\end{table}


\subsubsection{Performance Evaluation of the RS-GPT4V-Instruct Dataset}

    In evaluating different models on the RS-GPT4V-Instruct dataset, we focused on two core metrics: complex reasoning \& conversation and detailed description. These metrics aim to assess the models' ability to understand and generate complex dialogues and accurately capture details. As shown in Table \ref{table_4}, Full-FT performed the best in this evaluation, demonstrating its superior understanding and generation capabilities. Meanwhile, MoE-LoRA and LoRA also showed strong performance in these metrics. In contrast, other models such as Qwen-vl-Chat, LLaVA-1.5, and InternLM-XC2 scored relatively lower, indicating their potential limitations in handling complex reasoning, conversation, and detailed descriptions in remote sensing tasks.

\subsection{Limitations and Discussion}
\label{Limitations}

    \textbf{Limitations:} Although the RS-GPT4V dataset integrates multiple data sources and covers a wide range of remote sensing scenarios and target categories, its overall scale remains limited, particularly in specific application scenarios involving infrared and SAR modalities, where data support is insufficient. Despite the use of manual correction and advanced model generation to ensure data accuracy and high quality, issues of annotation errors and inconsistencies persist in large-scale datasets. Additionally, the models perform poorly in target detection tasks involving complex RSIs, especially when detecting small targets or targets within highly complex backgrounds.
    
    \textbf{Societal Impact:} Training large-scale multimodal language models requires significant computational resources, which can lead to significant energy consumption and corresponding carbon emissions. In the long run, this high energy consumption model may put continuous pressure on the environment.


\section{Conclusion}

    This study successfully constructed and thoroughly analyzed the RS-GPT4V dataset, a unified and multifunctional multimodal instruction-following dataset designed for remote sensing visual language tasks. By integrating a wide range of remote sensing scenarios and target categories, the RS-GPT4V dataset supports various tasks such as image description, visual question answering, complex scene understanding, visual reasoning, and task planning, using a unified (Question, Answer) format, significantly enhancing the dataset's practicality. During its construction, we adhered to key design principles of unity, diversity, correctness, richness, complexity, and robustness, ensuring the dataset's high quality and broad applicability to different visual and visual language tasks. Our experimental results demonstrate that fine-tuning with the RS-GPT4V dataset significantly improves the performance and generalization capabilities of multimodal large language models in executing complex task instructions. Overall, the release of the RS-GPT4V dataset provides strong support for remote sensing visual language research, and its design and implementation example will help advance remote sensing technology in various fields. In the future, we plan to expand the dataset to include other remote sensing domains, such as infrared and SAR modalities, further enhancing the models' adaptability.

\clearpage
\bibliographystyle{unsrt}

\medskip

{
\small
\bibliography{neurips}

\begin{thebibliography}{10}

\bibitem{xia2017aid}
Gui-Song Xia, Jingwen Hu, Fan Hu, Baoguang Shi, Xiang Bai, Yanfei Zhong, Liangpei Zhang, and Xiaoqiang Lu.
\newblock Aid: A benchmark data set for performance evaluation of aerial scene classification.
\newblock {\em IEEE Transactions on Geoscience and Remote Sensing}, 55(7):3965--3981, 2017.

\bibitem{ren2015faster}
Shaoqing Ren, Kaiming He, Ross Girshick, and Jian Sun.
\newblock Faster r-cnn: Towards real-time object detection with region proposal networks.
\newblock {\em Advances in neural information processing systems}, 28, 2015.

\bibitem{karpathy2015deep}
Andrej Karpathy and Li~Fei-Fei.
\newblock Deep visual-semantic alignments for generating image descriptions.
\newblock In {\em Proceedings of the IEEE conference on computer vision and pattern recognition}, pages 3128--3137, 2015.

\bibitem{li2019visualbert}
Liunian~Harold Li, Mark Yatskar, Da~Yin, Cho-Jui Hsieh, and Kai-Wei Chang.
\newblock Visualbert: A simple and performant baseline for vision and language.
\newblock {\em arXiv preprint arXiv:1908.03557}, 2019.

\bibitem{achiam2023gpt}
Josh Achiam, Steven Adler, Sandhini Agarwal, Lama Ahmad, Ilge Akkaya, Florencia~Leoni Aleman, Diogo Almeida, Janko Altenschmidt, Sam Altman, Shyamal Anadkat, et~al.
\newblock Gpt-4 technical report.
\newblock {\em arXiv preprint arXiv:2303.08774}, 2023.

\bibitem{touvron2023llama}
Hugo Touvron, Louis Martin, Kevin Stone, Peter Albert, Amjad Almahairi, Yasmine Babaei, Nikolay Bashlykov, Soumya Batra, Prajjwal Bhargava, Shruti Bhosale, et~al.
\newblock Llama 2: Open foundation and fine-tuned chat models.
\newblock {\em arXiv preprint arXiv:2307.09288}, 2023.

\bibitem{wang2023samrs}
Di~Wang, Jing Zhang, Bo~Du, Minqiang Xu, Lin Liu, Dacheng Tao, and Liangpei Zhang.
\newblock Samrs: Scaling-up remote sensing segmentation dataset with segment anything model, 2023.

\bibitem{sun2021fair1m}
Xian Sun, Peijin Wang, Zhiyuan Yan, Feng Xu, Ruiping Wang, Wenhui Diao, Jin Chen, Jihao Li, Yingchao Feng, Tao Xu, Martin Weinmann, Stefan Hinz, Cheng Wang, and Kun Fu.
\newblock Fair1m: A benchmark dataset for fine-grained object recognition in high-resolution remote sensing imagery, 2021.

\bibitem{ji2024changenet}
Deyi Ji, Siqi Gao, Mingyuan Tao, Hongtao Lu, and Feng Zhao.
\newblock Changenet: Multi-temporal asymmetric change detection dataset, 2024.

\bibitem{radford2021learning}
Alec Radford, Jong~Wook Kim, Chris Hallacy, Aditya Ramesh, Gabriel Goh, Sandhini Agarwal, Girish Sastry, Amanda Askell, Pamela Mishkin, Jack Clark, Gretchen Krueger, and Ilya Sutskever.
\newblock Learning transferable visual models from natural language supervision, 2021.

\bibitem{Long2021DiRS}
Yang Long, Gui-Song Xia, Shengyang Li, Wen Yang, Michael~Ying Yang, Xiao~Xiang Zhu, Liangpei Zhang, and Deren Li.
\newblock On creating benchmark dataset for aerial image interpretation: Reviews, guidances and million-aid.
\newblock {\em IEEE Journal of Selected Topics in Applied Earth Observations and Remote Sensing}, 14:4205--4230, 2021.

\bibitem{wang2023skyscript}
Zhecheng Wang, Rajanie Prabha, Tianyuan Huang, Jiajun Wu, and Ram Rajagopal.
\newblock Skyscript: A large and semantically diverse vision-language dataset for remote sensing, 2023.

\bibitem{muhtar2024lhrsbot}
Dilxat Muhtar, Zhenshi Li, Feng Gu, Xueliang Zhang, and Pengfeng Xiao.
\newblock Lhrs-bot: Empowering remote sensing with vgi-enhanced large multimodal language model, 2024.

\bibitem{huang2023visual}
Jiaxing Huang, Jingyi Zhang, Kai Jiang, Han Qiu, and Shijian Lu.
\newblock Visual instruction tuning towards general-purpose multimodal model: A survey.
\newblock {\em arXiv preprint arXiv:2312.16602}, 2023.

\bibitem{zhang2024earthgpt}
Wei Zhang, Miaoxin Cai, Tong Zhang, Yin Zhuang, and Xuerui Mao.
\newblock Earthgpt: A universal multi-modal large language model for multi-sensor image comprehension in remote sensing domain.
\newblock {\em arXiv preprint arXiv:2401.16822}, 2024.

\bibitem{rs16091477}
Yakoub Bazi, Laila Bashmal, Mohamad~Mahmoud Al~Rahhal, Riccardo Ricci, and Farid Melgani.
\newblock Rs-llava: A large vision-language model for joint captioning and question answering in remote sensing imagery.
\newblock {\em Remote Sensing}, 16(9), 2024.

\bibitem{pang2024h2rsvlm}
Chao Pang, Jiang Wu, Jiayu Li, Yi~Liu, Jiaxing Sun, Weijia Li, Xingxing Weng, Shuai Wang, Litong Feng, Gui-Song Xia, and Conghui He.
\newblock H2rsvlm: Towards helpful and honest remote sensing large vision language model, 2024.

\bibitem{qu2016deep}
Bo~Qu, Xuelong Li, Dacheng Tao, and Xiaoqiang Lu.
\newblock Deep semantic understanding of high resolution remote sensing image.
\newblock In {\em 2016 International conference on computer, information and telecommunication systems (Cits)}, pages 1--5. IEEE, 2016.

\bibitem{lobry2020rsvqa}
Sylvain Lobry, Diego Marcos, Jesse Murray, and Devis Tuia.
\newblock Rsvqa: Visual question answering for remote sensing data.
\newblock {\em IEEE Transactions on Geoscience and Remote Sensing}, 58(12):8555--8566, 2020.

\bibitem{Li_2020}
Ke~Li, Gang Wan, Gong Cheng, Liqiu Meng, and Junwei Han.
\newblock Object detection in optical remote sensing images: A survey and a new benchmark.
\newblock {\em ISPRS Journal of Photogrammetry and Remote Sensing}, 159:296–307, January 2020.

\bibitem{kuckreja2023geochat}
Kartik Kuckreja, Muhammad~Sohail Danish, Muzammal Naseer, Abhijit Das, Salman Khan, and Fahad~Shahbaz Khan.
\newblock Geochat: Grounded large vision-language model for remote sensing.
\newblock {\em arXiv preprint arXiv:2311.15826}, 2023.

\bibitem{zhang2023instruction}
Shengyu Zhang, Linfeng Dong, Xiaoya Li, Sen Zhang, Xiaofei Sun, Shuhe Wang, Jiwei Li, Runyi Hu, Tianwei Zhang, Fei Wu, et~al.
\newblock Instruction tuning for large language models: A survey.
\newblock {\em arXiv preprint arXiv:2308.10792}, 2023.

\bibitem{li2023vision}
Chen Li, Yixiao Ge, Dian Li, and Ying Shan.
\newblock Vision-language instruction tuning: A review and analysis.
\newblock {\em arXiv preprint arXiv:2311.08172}, 2023.

\bibitem{you2023ferret}
Haoxuan You, Haotian Zhang, Zhe Gan, Xianzhi Du, Bowen Zhang, Zirui Wang, Liangliang Cao, Shih-Fu Chang, and Yinfei Yang.
\newblock Ferret: Refer and ground anything anywhere at any granularity.
\newblock {\em arXiv preprint arXiv:2310.07704}, 2023.

\bibitem{bsharat2023principled}
Sondos~Mahmoud Bsharat, Aidar Myrzakhan, and Zhiqiang Shen.
\newblock Principled instructions are all you need for questioning llama-1/2, gpt-3.5/4.
\newblock {\em arXiv preprint arXiv:2312.16171}, 2023.

\bibitem{cheng2022nwpu}
Qimin Cheng, Haiyan Huang, Yuan Xu, Yuzhuo Zhou, Huanying Li, and Zhongyuan Wang.
\newblock Nwpu-captions dataset and mlca-net for remote sensing image captioning.
\newblock {\em IEEE Transactions on Geoscience and Remote Sensing}, 60:1--19, 2022.

\bibitem{lu2017exploring}
Xiaoqiang Lu, Binqiang Wang, Xiangtao Zheng, and Xuelong Li.
\newblock Exploring models and data for remote sensing image caption generation.
\newblock {\em IEEE Transactions on Geoscience and Remote Sensing}, 56(4):2183--2195, 2017.

\bibitem{RSITMD}
Zhiqiang Yuan, Wenkai Zhang, Kun Fu, Xuan Li, Chubo Deng, Hongqi Wang, and Xian Sun.
\newblock Exploring a fine-grained multiscale method for cross-modal remote sensing image retrieval.
\newblock {\em IEEE Transactions on Geoscience and Remote Sensing}, 60:1–19, 2022.

\bibitem{rahnemoonfar2021floodnet}
Maryam Rahnemoonfar, Tashnim Chowdhury, Argho Sarkar, Debvrat Varshney, Masoud Yari, and Robin~Roberson Murphy.
\newblock Floodnet: A high resolution aerial imagery dataset for post flood scene understanding.
\newblock {\em IEEE Access}, 9:89644--89654, 2021.

\bibitem{RSIVQA}
Sylvain Lobry, Diego Marcos, Jesse Murray, and Devis Tuia.
\newblock Rsvqa: Visual question answering for remote sensing data.
\newblock {\em IEEE Transactions on Geoscience and Remote Sensing}, 58(12):8555–8566, December 2020.

\bibitem{zhan2023rsvg}
Yang Zhan, Zhitong Xiong, and Yuan Yuan.
\newblock Rsvg: Exploring data and models for visual grounding on remote sensing data.
\newblock {\em IEEE Transactions on Geoscience and Remote Sensing}, 61:1--13, 2023.

\bibitem{liu2024visual}
Haotian Liu, Chunyuan Li, Qingyang Wu, and Yong~Jae Lee.
\newblock Visual instruction tuning.
\newblock {\em Advances in neural information processing systems}, 36, 2024.

\bibitem{hu2021lora}
Edward~J. Hu, Yelong Shen, Phillip Wallis, Zeyuan Allen-Zhu, Yuanzhi Li, Shean Wang, Lu~Wang, and Weizhu Chen.
\newblock Lora: Low-rank adaptation of large language models, 2021.

\bibitem{liu2023moelora}
Qidong Liu, Xian Wu, Xiangyu Zhao, Yuanshao Zhu, Derong Xu, Feng Tian, and Yefeng Zheng.
\newblock Moelora: An moe-based parameter efficient fine-tuning method for multi-task medical applications.
\newblock {\em arXiv preprint arXiv:2310.18339}, 2023.

\bibitem{papineni2002bleu}
Kishore Papineni, Salim Roukos, Todd Ward, and Wei-Jing Zhu.
\newblock Bleu: a method for automatic evaluation of machine translation.
\newblock In {\em Proceedings of the 40th annual meeting of the Association for Computational Linguistics}, pages 311--318, 2002.

\bibitem{banerjee-lavie-2005-meteor}
Satanjeev Banerjee and Alon Lavie.
\newblock {METEOR}: An automatic metric for {MT} evaluation with improved correlation with human judgments.
\newblock In Jade Goldstein, Alon Lavie, Chin-Yew Lin, and Clare Voss, editors, {\em Proceedings of the {ACL} Workshop on Intrinsic and Extrinsic Evaluation Measures for Machine Translation and/or Summarization}, pages 65--72, Ann Arbor, Michigan, June 2005. Association for Computational Linguistics.

\bibitem{ROUGE_L}
Chin-Yew Lin.
\newblock {ROUGE}: A package for automatic evaluation of summaries.
\newblock In {\em Text Summarization Branches Out}, pages 74--81, Barcelona, Spain, July 2004. Association for Computational Linguistics.

\bibitem{cider}
Ramakrishna Vedantam, C.~Lawrence Zitnick, and Devi Parikh.
\newblock Cider: Consensus-based image description evaluation, 2015.

\bibitem{anderson2016spice}
Peter Anderson, Basura Fernando, Mark Johnson, and Stephen Gould.
\newblock Spice: Semantic propositional image caption evaluation, 2016.

\bibitem{bazi2022bi}
Yakoub Bazi, Mohamad~Mahmoud Al~Rahhal, Mohamed~Lamine Mekhalfi, Mansour~Abdulaziz Al~Zuair, and Farid Melgani.
\newblock Bi-modal transformer-based approach for visual question answering in remote sensing imagery.
\newblock {\em IEEE Transactions on Geoscience and Remote Sensing}, 60:1--11, 2022.

\bibitem{zhang2023spatial}
Zixiao Zhang, Licheng Jiao, Lingling Li, Xu~Liu, Puhua Chen, Fang Liu, Yuxuan Li, and Zhicheng Guo.
\newblock A spatial hierarchical reasoning network for remote sensing visual question answering.
\newblock {\em IEEE Transactions on Geoscience and Remote Sensing}, 61:1--15, 2023.

\bibitem{bai2023qwen}
Jinze Bai, Shuai Bai, Shusheng Yang, Shijie Wang, Sinan Tan, Peng Wang, Junyang Lin, Chang Zhou, and Jingren Zhou.
\newblock Qwen-vl: A frontier large vision-language model with versatile abilities.
\newblock {\em arXiv preprint arXiv:2308.12966}, 2023.

\bibitem{liu2023improved}
Haotian Liu, Chunyuan Li, Yuheng Li, and Yong~Jae Lee.
\newblock Improved baselines with visual instruction tuning.
\newblock {\em arXiv preprint arXiv:2310.03744}, 2023.

\end{thebibliography}

}

\clearpage
\appendix

\section{Model Architecture}

    To comprehensively evaluate the performance of the RS-GPT4V dataset, we used LLaVA-1.5-7B as the pre-trained model. This model is pre-trained using rich visual language data to bridge the gap between vision and language, with the trainable parameters of the model denoted as $\theta$. Specifically, \(X_v\) represents the input image, \(X_{instruct}\) represents the text instruction, \(L\) is the sequence length of the answer \(X_a\), and \(X_{a,<i}\) represents all answer tokens before the current prediction token \(x_i\), where \(i\) denotes the steps during text token generation. We calculate the probability of the entire target answer \(X_a\) as follows:

\vspace{-0.5cm} 
{\setlength\abovedisplayskip{0cm}
\setlength\belowdisplayskip{0.3cm}
\begin{equation}
p\left(X_a|X_v,X_{instruct}\right)=\prod_{i=1}^{L} p_\theta\left(x_i|X_v,X_{instruct,<i},X_{a,<i}\right)
\end{equation}}

\begin{figure}[ht]
  \centering
  \includegraphics[width=1\textwidth]{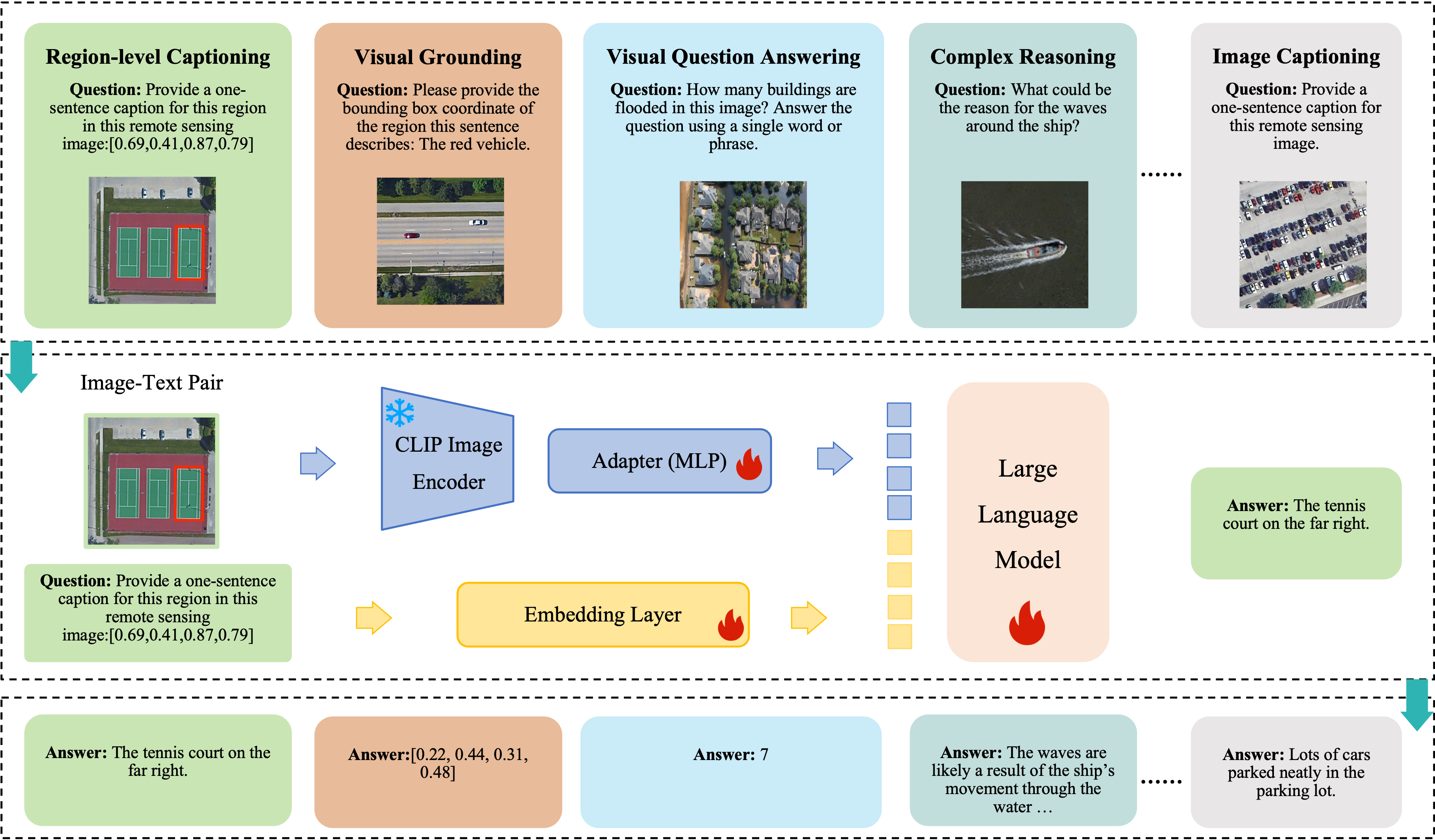}
  \caption{\textbf{Model Architecture for RS-GPT4V Supervised Fine-Tuning.} Illustrates the input of image-text pairs and the resulting outputs when fine-tuning with the RS-GPT4V dataset.}
  \label{figure_4}
\end{figure}

    As shown in the Figure \ref{figure_4}, LLaVA-1.5-7B employs CLIP-336px to retain more information from the original pixel space and uses Vicuna v1.5 as the language encoder. Additionally, the model's multimodal connector adopts a two-layer MLP structure. During training, the image encoder first extracts visual tokens from the input image. Then, the encoder encodes the image into a series of image tokens. Next, the generated token sequence is passed through a two-layer MLP, which uses the GELU activation function to map the visual tokens to the embedding space dimension. The mapped image features are combined with the text instruction tokens to form the input for the large language model. The architecture design of LLaVA-1.5-7B enables it to effectively handle various remote sensing tasks, including region-level descriptions, visual localization, visual question answering, complex reasoning, and image descriptions. Through comprehensive training on these tasks, the model can accurately identify objects and scenes in images and perform complex reasoning and descriptions.

\section{RS-GPT4V-Instruct}
   We conducted a comprehensive comparison of the RS-GPT4V-Instruct dataset, generated through Instruction-Response Generation, with several other remote sensing instruction datasets. As shown in Table \ref{tab:2}, all instruction datasets support complex reasoning and multi-turn dialogues. However, the RS-GPT4V-Instruct dataset undergoes final corrections through manual verification, significantly enhancing its accuracy. Other datasets, such as GeoChat, LHRS-Instruct, and HqDC-Instruct, primarily adopt a two-stage process: the first stage uses multimodal language models (MLLMs) to generate detailed descriptions and extract fine-grained information, while the second stage employs large language models (LLMs) for Instruction-Response Generation, complex reasoning, and multi-turn dialogues. Errors in the initial annotations tend to accumulate in subsequent processing. Due to the lack of visual signal support in the second stage, the generated multi-turn dialogues are limited to the annotation information from the first stage. In contrast, the RS-GPT4V-Instruct dataset integrates visual signals, capturing object information and image background details beyond the system prompts, effectively reducing the accumulation of annotation errors. This approach ensures higher accuracy, making the RS-GPT4V-Instruct dataset perform exceptionally well in tasks involving image content.

\begin{table}[ht]
\vspace{-2mm}
    \caption{\textbf{Comparative of RS-GPT4V-Instruct and Other Remote Sensing Multimodal Instruction-Following Datasets.}}   
    \scriptsize 
    \label{tab:2}
    \resizebox{\textwidth}{!}{%
        \begin{tabular}{ccccc} 
            \toprule
            Instruction-Following Dataset & GeoChat & LHRS-Instruct & HqDC-Instruct & \textbf{RS-GPT4V-Instruct} \\
            \midrule
            Data Source & 
            \begin{tabular}[c]{@{}c@{}}DOTA\\ DIOR\\ FAIR1M\end{tabular} & 
            \begin{tabular}[c]{@{}c@{}}RSITMD\\ NWPU-Captions\\ LHRS-Align\end{tabular} & 
            \begin{tabular}[c]{@{}c@{}}DOTA\\ FAIR1M\end{tabular} & 
            DIOR \\
            Labelling Model & Vicuna-v1.5 & 
            \begin{tabular}[c]{@{}c@{}}Vicuna-v1.5\\ \& GPT-4\end{tabular} & 
            language-only Gemini & 
            GPT-4V \\
            Visual Signal Accessibility & \ding{55} & \ding{55} & \ding{55} & \ding{51} \\
            Complex Reasoning & \ding{51} & \ding{51} & \ding{51} & \ding{51} \\
            High Quality & \ding{55} & \ding{51} & \ding{55} & \ding{51} \\
            Multi-turn Conversation & \ding{51} & \ding{51} & \ding{51} & \ding{51} \\
            \midrule
        \end{tabular}%
        
    }
    \vspace{-2mm}
\end{table}

\end{document}